\documentclass[journal]{IEEEtran}
\usepackage{cite}
\usepackage{times}
\usepackage{epsfig}
\usepackage{graphicx}
\usepackage{amsmath}
\usepackage{amssymb}
\usepackage{multirow}
\usepackage{tabularx}
\usepackage{hyperref}
\usepackage{booktabs}
\usepackage{array}
\usepackage{amsfonts}
\usepackage{xcolor}
\newcolumntype{L}{>{$}l<{$}}
\newcolumntype{C}{>{$}c<{$}}
\newcolumntype{R}{>{$}r<{$}}
\newcolumntype{P}[1]{>{\centering\arraybackslash}p{#1}}

\begin{document}
\title{Structure-Aware Feature Generation for Zero-Shot Learning} \author{Lianbo
  Zhang, Shaoli Huang,~\IEEEmembership{Member,~IEEE}, Xinchao
  Wang,~\IEEEmembership{Senior Member, IEEE}, Wei
  Liu,~\IEEEmembership{Senior Member,~IEEE}, and Dacheng
  Tao,~\IEEEmembership{Fellow, IEEE}
  \thanks{Lianbo Zhang and Wei Liu are with School of Computer Science, FEIT,
    University of Technology Sydney, NSW 2007, Australia (e-mail:
    lianbo.zhang@student.uts.edu.au;
    wei.liu@uts.edu.au).}
  \thanks{Shaoli Huang is with School of Computer
    Science, FEIT, University of Sydney, NSW 2008, Australia (email:
    shaoli.huang@sydney.edu.au).} \thanks{Xinchao Wang is with Department of
    Electrical and Computer Enigneering, National University of Singapore
    (email:xinchao@nus.edu.sg).} \thanks{Dacheng Tao is with the JD Explore
    Academy, JD.com, China (e-mail: dacheng.tao@gmail.com).}
}


\maketitle

\begin{abstract}
  Zero-Shot Learning (ZSL) targets at recognizing unseen categories by
  leveraging auxiliary information, such as attribute embedding. Despite the
  encouraging results achieved, prior ZSL approaches focus on improving the
  discriminant power of seen-class features, yet have largely overlooked the
  geometric structure of the samples and the prototypes. The subsequent
  attribute-based generative adversarial network (GAN), as a result, also
  neglects the topological information in sample generation and further yields
  inferior performances in classifying the visual features of unseen classes. In
  this paper, we introduce a novel structure-aware feature generation scheme,
  termed as SA-GAN, to explicitly account for the topological structure in
  learning both the latent space and the generative networks. Specifically, we
  introduce a constraint loss to preserve the initial geometric structure when
  learning a discriminative latent space, and carry out our GAN training with
  additional supervising signals from a structure-aware discriminator and a
  reconstruction module. The former supervision distinguishes fake and real
  samples based on their affinity to class prototypes, while the latter aims to
  reconstruct the original feature space from the generated latent space. This
  topology-preserving mechanism enables our method to significantly enhance the
  generalization capability on unseen-classes and consequently improve the
  classification performance. Experiments on four benchmarks demonstrate that
  the proposed approach consistently outperforms the state of the art. Our code
  can be found in the supplementary material and will also be made publicly
  available.
\end{abstract}


%
\IEEEpeerreviewmaketitle

\section{Introduction}
Zero-Shot Learning~(ZSL) strives to recognize samples of unseen classes given
their semantic descriptions, and have in recently years emerged as a
widely-studied task due to their practical nature~\cite{palatucci2009zero,
  larochelle2008zero, lampert2009learning}. Typical ZSL
approaches~\cite{lampert2013attribute, frome2013devise, zhang2017learning}
tackles the problem by directly finding a shared latent space that aligns visual
features and semantic embedding. However, learning such a shared space from the
seen data have been found highly challenging, since the gap between the visual
and semantic space is in many cases significant~\cite{long2017zero}.

\begin{figure}[h]
  \begin{center}
    \includegraphics[width=1.0\linewidth]{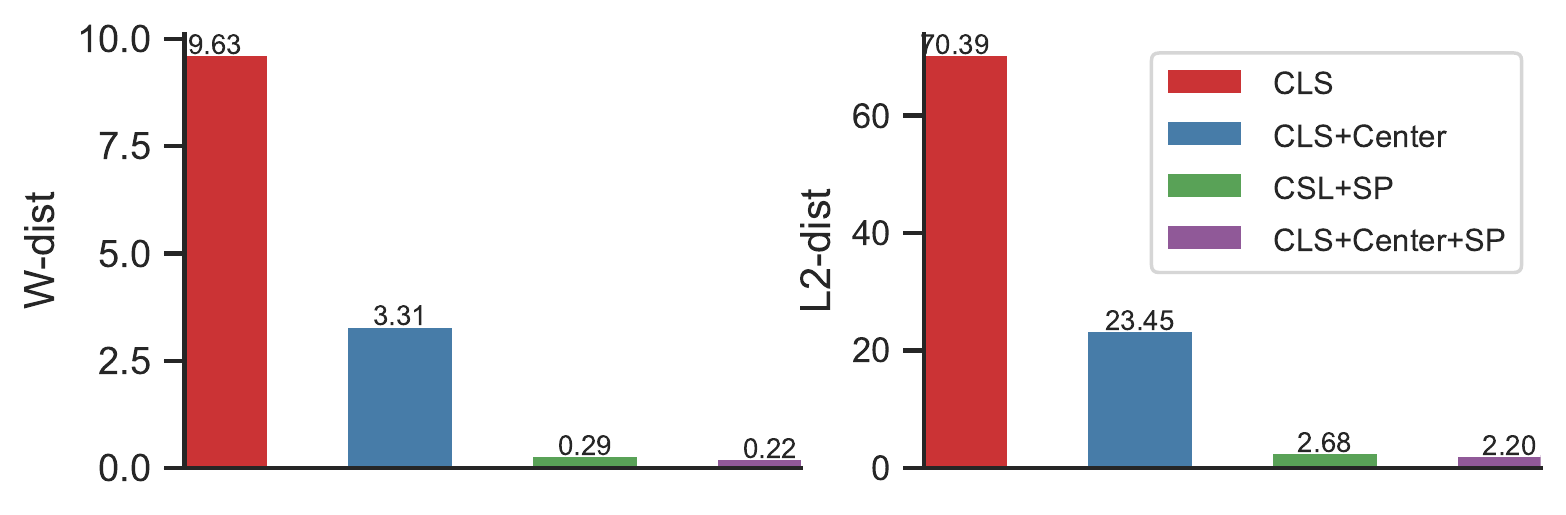}
  \end{center}
\vspace{-1em}
\caption{We quantitatively measure the average change of feature-prototype
  distance between the original visual space and the latent space on CUB dataste
  \cite{wah2011caltech}. W-dist and L2-dist respectively denote Wasserstein
  distance and Euclidean distance. CLS denotes the classification loss, Center
  denotes the center constraint, and SP denotes the introduced
  structure-preserving constraint. A higher value indicates a greater change in
  geometric structure.}
  \label{fig:structure-change}
  \vspace{-1.0em}
\end{figure}

\begin{figure*}[h]
  \begin{center}
    \includegraphics[width=1.0\linewidth]{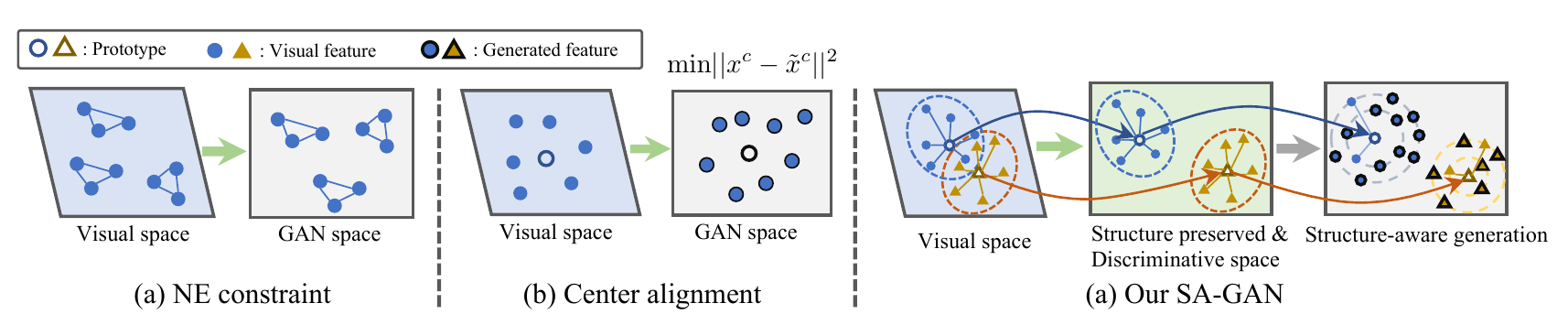} \\
  \end{center}
  \caption{Comparison of different visual structure constraints for feature
    generation. (a) NE constraint \cite{tran2019improving} aims to maintain the
    neighbourhood structure between the visual and GAN space. (b) Center
    alignment \cite{wan2019transductive} clusters fake samples to find visual
    centers and align the fake centers with the that of real ones. Here, $x^c$
    and $\tilde{x}^c$ denote the class centers. (c) The proposed SA-GAN.
    Compared with existing methods, besides the difference in structure
    definition, our approach can better maintain the original structure
    information by using the mapped rather than the newly calculated prototype
    as a reference. Moreover, our method incorporates the prototype as condition
    input into the discriminator, which is more effective than adding a
    constraint loss to enforce the GAN to consider structure information. This
    is because the discriminator is usually the key to update the generator.}
  \label{fig:title}
  \vspace{-0.5em}
\end{figure*}

The works of~\cite{xian2018feature, felix2018multi, xian2019f} bypass learning
such a direct visual-semantic alignment and resort to an alternative paradigm
based on sample generation. They follow the workflow of first learning a sample
generator conditioned on semantic embedding and then generating visual features
of unseen classes for training the ZSL classifier. More recent works
of~\cite{han2020learning, wan2019transductive, paul2019semantically} further
improve this learning pipeline by introducing a mapping stage to enhance the
feature quality, such as reducing redundancy \cite{han2020learning} or
increasing discriminability \cite{paul2019semantically}. Though
sample-generation-based approaches have demonstrated promising results, they
still suffer from the issue that the generated samples for unseen-class are
prone to collapsing to the seen-class centers.

In this paper, we look into the generalization capability of ZSL methods on
unseen data, by explicitly incorporating the topological structure, defined as
the geometry relationship between samples and their corresponding prototypes,
into the learning of the latent space and the generator. The rationale behind
this design lies in our experimental observation: without a topology-preserving
constraint, the data topological structure change abruptly throughout the
feature mapping, indicating that the model fails to capture the intrinsic data
distribution and potentially leads to the bias towards the seen categories. We
show some visual examples in Fig.~\ref{fig:structure-change}

Specifically, our proposed scheme comprises two stages, structure-preserving
mapping and structure-aware feature generation, as illustrated in
Fig.~\ref{fig:title}(c). In the first stage, we impose a structure-preserving
constraint in learning the mapping from visual features to the latent space.
This constraint function aims to increase features' discriminative power while
penalizing any deviation to the original prototype structure. In the second
stage, we enforce the GAN learning to handle the prototype structure, through
utilizing additional supervision signals from a structure-aware discriminator
and a feature reconstruction module. The discriminator is devised to distinguish
fake samples from real ones conditioned on the corresponding prototype, while
the reconstruction module ensures that the derived latent features may
reconstruct the original ones.

We evaluate the proposed method on four challenging benchmark datasets under
both the ZSL and the Generalized~ZSL~(GZSL) setting. Experimental results
demonstrate that our approach significantly outperforms existing methods on most
datasets under both settings. Our method achieves impressive performance on
AWA2, CUB, SUN, and FLO datasets with harmonic mean accuracies of $68.0\%$,
$58.7\%$, $44.4\%$ and $86.4\%$, respectively. Besides, using VAEGAN as a
baseline, SA-GAN improves the unseen class recognition by 2.3\%, 7.3\%, 6.2\%,
and 28.2\% on the above four datasets, validating the generalization capability
of our method.

Our contribution is therefore a novel SA-GAN scheme that explicitly accounts for
topological structure of samples throughout the ZSL pipeline, in which a
structure-preserving constraint is imposed into the feature mapping and a
structure-aware discriminator is introduced for the GAN to account for the
sample distribution. SA-GAN delivers gratifying results consistently superior to
the state of the art over four benchmarks, and significantly boosts the accuracy
over unseen classes.

\section{Related Works}
\begin{figure*}[h]
  \begin{center}
    \includegraphics[width=0.95\linewidth]{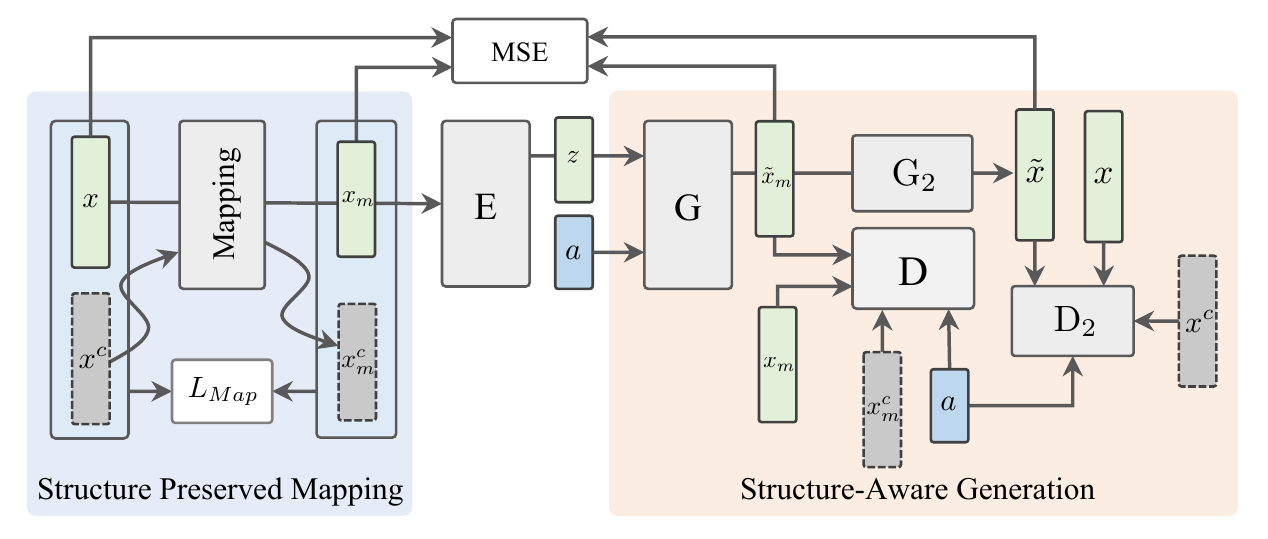}
  \end{center}
  \caption{The framework of the proposed GZSL method. The latent feature $x_m$
    is extracted by the mapping sub-network. The generator $G(\cdot)$
    synthesizes new features $\tilde{x}_m$ based on class embedding $a$ and
    random noise $z$. The discriminator $D(\cdot)$ tries to distinguish between
    real and fake instances by measuring the relationship with the class
    embedding $a$ and prototype $x^c_m$. The second generator $G_2$ tries to
    recover the original visual features from the synthetic latent ones.}
  \label{fig:framework}
\end{figure*}

\subsection{Zero-Shot Learning}
Zero-shot learning relies on the side information to exploit knowledge
transferring from seen classes to a disjoint set of unseen classes. The side
information can be class-level semantic descriptions or features, e.g., semantic
attributes \cite{akata2015evaluation, farhadi2009describing,
  parikh2011relative}, and word vectors \cite{mikolov2013efficient,
  mikolov2013distributed} for bridging the gap between disjoint seen and unseen
domain. Early researches focus on the conventional ZSL problem
\cite{norouzi2013zero, socher2013zero, frome2013devise, akata2015evaluation,
  zhang2015zero}, which is dominated by the semantic embedding approaches
\cite{bucher2016improving, frome2013devise, romera2015embarrassingly,
  kodirov2017semantic}. These methods learn to encode the visual features of
seen class into the semantic spaces. For example, Socher \textit{et al.}
\cite{socher2013zero} learns a deep non-linear mapping between images and tags.
Norouzi \textit{et al.} \cite{norouzi2013zero} uses the probabilities of a
softmax-output layer to weight the vectors of all the classes. Romera-Paredes
\textit{et al.} \cite{romera2015embarrassingly} proposes to learn
visual-semantic mapping by optimizing a simple objective function with a
closed-form solution. By doing so, the visual feature and the semantic embedding
will share the same space, and the ZSL recognition can be accomplished by
adopting the nearest neighbor classifier to assign given instances to
corresponding classes or maximizing the compatibility score using various
distance metrics. A well-trained ZSL model is then expected to maintain this
ability in the unseen domain. Some works \cite{akata2013label,
  akata2015evaluation, frome2013devise, kodirov2017semantic} also attempt to
learn an inverse mapping from the semantic vectors to the visual one. For
example, DeViSE \cite{frome2013devise} learn a bilinear compatibility function
using a multiclass with an efficient ranking formulation. Kodirov
\cite{kodirov2017semantic} uses an encoder to project the visual features into a
semantic space and a decode to reconstruct the original visual space to preserve
more information contained in the visual space.

Conventional ZSL tests only on the unseen classes, while the generalized
zero-shot learning \cite{zhang2018triple, xian2018zero, rahman2018unified,
  li2020joint, jia2019deep} is sparking more interest, where both seen and unseen classes are
available in the inference stage. Since GZSL recognize the images from both seen
and unseen classes, it is a more challenging task and suffers from a more
serious data imbalance issue. Semantic embedding was developed for conventional
ZSL fail to resolve this problem in GZSL. They tend to be highly overfitting the
seen classes and harm the classification of unseen classes. This is verified by
Xian \textit{et al.} \cite{xian2018zero}, where they conducted experiments and
illustrated that almost all approaches designed for conventional ZSL see a
significant drop in the GZSL setting.

Generative ZSL methods \cite{xian2018feature, gao2020zero,
  schonfeld2019generalized, han2020learning, paul2019semantically, shen2019scalable} learn
Generative Adversarial Networks (GANs) \cite{goodfellow2014generative,
  felix2018multi, paul2019semantically, sariyildiz2019gradient} to synthesize
unseen class features, thereby converting the ZSL problem to a supervised
problem, then a ZSL classifier is learned for unseen class recognition. Among
them, a typical generative method, f-CLSWGAN \cite{xian2018feature}, uses
Wasserstein GAN (WGAN) \cite{arjovsky2017wasserstein} to learn the generator for
feature generation, which is achieved by optimizing a WGAN loss and a
classification loss to ensure the feature discriminative power. The classifier
can be replaced by an attribute regressor to guarantee the multi-modal
cycle-consistency \cite{felix2018multi}. Another generative method, CADA-VAE
\cite{schonfeld2019generalized} use two Variational Autoencoders (VAE)
\cite{kingma2013auto,doersch2016tutorial} to align the visual feature and class
embeddings in a shared latent space. By combining VAE and WGAN, Xian \textit{et
  al.} \cite{xian2019f} proposes f-VAEGAN, where GAN's generator is used as a
decoder of the VAE for feature generation. f-VAEGAN requires the decoder to
reconstruct the visual feature, thereby promoting the cycle-consistency between
generated and original visual features. Very recent works \cite{han2020learning,
  wan2019transductive, paul2019semantically} seeks a two-stage pipeline, where
the first stage projecting the visual feature to a latent space. The second
stage follows the standard generative methods. These methods improve feature
generation by integrating extra constraints to obtain a more appropriate latent
space. For example, SARB \cite{ paul2019semantically} fine-tunes a pre-trained
model with class embeddings to obtain a latent space that is discriminative and
hubness-free. RFF-GZSL \cite{han2020learning} restricts the mutual information
between original visual features and the latest features so that the redundant
information of original visual features can be removed without losing the
discriminative information. Despite promising results demonstrated by these
methods, they fail to consider the visual structure in learning the mapping and
feature generator.

\subsection{Structure-Aware GAN.} There have been works \cite{peng2016deep,
  tran2018dist, tran2019improving} in the literature considering structure
constraints in learning GAN networks. For example, PARTY \cite{peng2016deep}
embeds local structure and prior sparsity information into the hidden
representation learning to preserve the sparse reconstruction relation. Dist-GAN
\cite{tran2018dist} uses latent-data distance constraint to enforce the
compatibility between latent sample distance and the corresponding data sample
distances, which alleviates the model collapse. GN-GAN \cite{tran2019improving}
explores reversed t-SNE \cite{maaten2008visualizing} to regularize the GAN
training at both the local level and global level. In the context of ZSL, MCA
\cite{wan2019transductive} aligns semantic centers and visual centers in the
source domain data and imposes this structure to the target domain data. Other
works explore intra-class relations between prototypes and class samples, such
as applying center loss \cite{han2020learning, wen2016discriminative} to reduce
intra-class variance and improve class discrimination. Wan \textit{et al.}
\cite{wan2019transductive} also considers various metrics to align class centers
between visual space and GAN space.

In this paper, we improve the GAN-based ZSL method by using a prototype
structure. Our method is different from previous approaches as we consider the
structure constraint in both mapping and feature generation. To the best of our
knowledge, our method is the first to incorporate this kind of information in
the two-stage ZSL recognition context. We systematically design a
structure-preserving mapping module and structure-aware generation module that
generates discriminative instances carrying structure information.

\section{Method}
As shown in Fig.~\ref{fig:framework}, our proposed learning framework mainly
consists of two training stages: Structure-preserved mapping and structure-aware
generation. The first stage aims to learn a mapping function that increases the
feature discriminability while preserving the topological structure of data.
After obtaining the mapped data from the previous step, we train a
structural-aware generator using a discriminator incorporated with a prototype
condition.

\subsection{Notation}
In zero-shot learning, there is a set of images $X = \{x_1, \cdots , x_l \} \cup
\{x_{l+1}, \cdots, x_t \} $ encoded in the visual space $\mathcal{X}$, a seen
class label set $Y^s$, a unseen label set $Y^u$ and a class embedding set
$A={a(y)|y\in{Y^s \cup Y^u}}$. $A$ is encoded in the semantic embedding space
$\mathcal{A}$ that define the shared semantic relationship among categories. For
example, two categories can be described by different combinations of a set of
attributes. The first $l$ instances $x_s(x \leq l)$ are labeled as one of the
seen classes $y_s\in Y^s$ while the other instances $x_u(l+1\leq n \leq t)$
belong to the unseen classes $y_u \in Y^u$. Using training set that contains
only seen classes, the task of zero-shot learning is to predict the label of
those unlabeled instances of unseen classes, i.e. $f_{zsl}\rightarrow
\mathcal{Y}^u$. In generalized zero-shot learning, those unlabeled instances can
be either from seen or unseen classes, i.e. $f_{gzsl}\rightarrow \mathcal{Y}^s
\cup \mathcal{Y}^u$.

\subsection{Structure-Preserving Mapping}

\subsubsection{Discriminative Mapping} In practice, the mapping function can be
learned under the supervision of a classifier to ensure the discriminative power
of the latent feature with the objective formulated as

\begin{equation}
  L_{cls} = - \frac{1}{N_s}\sum^{N_s}_{i=1}H(y_i, f_c({f_m({x_i})})
\end{equation}
where $H$ is the cross entropy loss betwen true and predicted label of seen class instance $x_i$.

To further enhance the feature discriminativity we also use the center loss
constraint which is proposde by Wen \textit{et al.}
\cite{wen2016discriminative},
\begin{equation}
  L_{center} = \frac{1}{N_s} \sum^m_{i=1}||f_{m}(x_i,m) - x^c_{m, y_i}||^2_2.
\end{equation}
Among them, $x^c_{m, y_i}$ denotes the $y_i$th class center. It is initialized as a parameter to be updated in a
mini-batch, and is jointly supervised by $L_{cls}$ and $L_{center}$. Together with classifier, the center loss greatly
enhances the discriminative power of the learned feature $x_m$.

\subsubsection{Structure-Preserving Constraint} As discussed previously, directly
learning a discriminative mapping will lead to losing the data's structural
information. In this section, we describe our proposed structural constraint to
address this issue. We first define the structural information as the geometry
relationship between samples and their corresponding prototypes. One naive way
to measure the structural change after the mapping is to compute the class
prototype separately in both original and mapped spaces, then compare the
difference of each sample-prototype distance in both spaces. However,
calculating the class prototype in each iteration is time-consuming make it
infeasible in practice. Moreover, the new computed prototype (by averaging the
same class's mapped features) is usually different from the mapped prototype (by
mapping the original prototype into the new space). To address these issues, we
propose to directly use the mapped prototype to characterize the structural
information in the new space. Therefore, our proposed structure-preserving
constraint $L_{sp}$that penalizes the relative structure changes is expressed
as:
\begin{equation}
  L_{sp} = \frac{1}{N_s}\sum^{N_s}_{i}\Big[\lVert {x_i - x^c_{y_i}} \rVert_2 - \lVert {f_{m}(x_i) - f_{m}(x^c_{y_i})}\rVert_2 \Big]^2
\end{equation}
where the prototype, $x^c_{y_i} \in \mathbb{R}^{d_s}$, is the mean vector of the
support instances sampled from the same class. The constant $l_i = \lVert {x_i -
  x^c_{y_i}} \rVert_2$ measures the $L_2$ distance between $x_i$ and its
prototype $x^c_{y_i}$. $f_{m}$ is the mapping function. We simultaneously map
the visual feature and its prototype onto the latent space and attempt to
preserve their geometric distance which can be defined using any distance
metrics such as Eucliden distance, cosine similariy, et al.

\subsubsection{Mapping Objective} The classifier, center-loss, and
structure-preserving constraint are learned simultaneously by joint minimizing
the loss function represented in equation 1, 2, and 3 weighted by the factor
$\gamma_c$ and $\gamma_s$.

\begin{equation}
  L_{Map} = L_{cls} + \gamma_{c} L_{center} + \gamma_{s} L_{sp},
\end{equation}
where $\gamma_{c}$ and $\gamma_{s}$ are set to 0.01 and 1.0 in this paper.
$L_{cls}$ and $L_{c}$ ensure that the latent feature is discriminative to
benefit category recognition. The former two focuses on separability of the seen
classes by the other information such as prototype structure are not considered.
This issue is addressed by the prototype structure constraint $L_{sp}$.

At the end of stage 1, the visual feature in the new space can implicitly encode
structural information within discriminative representations. Therefore, it not
only enjoyes discriminative power but also retain a high-quality structure. The
mapping function carrying these characteristics provides a well-generalized
space for subsequent GAN learning.

\subsection{Structure-Aware Feature Generation}
\subsubsection{Mapped GAN} In the second stage, we aim to encourage the generator to
produce discriminative and structure-aware features in latent space. We achieve
this by modifying the discriminator to include class prototypes as an auxiliary
input to train the conditional generator. It is then optimized by a Wasserstein
adversarial loss defined by

\begin{equation}
\begin{split}
  L_{mWGAN}  &= \mathbb{E}[D(x_m, a_{y}, x^c_{m,y}; \; \theta_D)] \\
           &- \mathbb{E}[D(\tilde{x}_m, a_{y}, x^c_{m,y}; \; \theta_D)] \\
           &- \lambda \mathbb{E}[(|| \triangledown_{\hat{x}_m}D(\hat{x}_m; \; \theta_D)||_2 - 1)^2] 
\end{split}
\end{equation}
where $G(\cdot | \theta_{G})$ is the generator, $D(\cdot | \theta_{D})$ denote the discriminator. The generator synthesizes instances $\tilde{x}_m = G(z, a_y)$ conditioning on class embedding $a$ and a multi-dimension Gaussian distribution $z \sim \mathcal{N}(0, 1)$. Besides, $\hat{x}_m=\alpha x_m + (1-\alpha)\tilde{x}_m$ where $\alpha \sim U(0, 1)$, and $\lambda$ is the penalty coefficient.

\subsubsection{Reconstructed GAN} Note that SA-GAN generates visual features in the structured-preserved latent space, but does not guarantee that the generated features are well fitted to the visual distribution, which might lead to ineffective learning of the final linear classifier. We conjecture that this issue could be alleviated by reconstructing the original feature space from the mapped space. To this end, we optimize a second GAN sub-network
\begin{equation}
  \begin{split}
    L_{rWGAN} &= \mathbb{E}[D_2(x, a_{y}, x^c_{y}; \; \theta_{D_2})] - \mathbb{E}[D_2(\tilde{x}, a_{y}, x^c_{y}; \; \theta_{D_2})] \\
    &- \lambda \mathbb{E}[(|| \triangledown_{\hat{x}}D_2(\hat{x}; \; \theta_{D_2})||_2 - 1)^2] 
  \end{split}
\end{equation}
where $\tilde{x} = G_2(x_m, a_y)$, $\hat{x}=\alpha x + (1-\alpha)\tilde{x}$ with $\alpha \sim U(0, 1)$, and $\lambda$ is the penalty coefficient. Minimizing the $L_{rWGAN}$ requires a high-quality recovery and rich information encoded in the generated features. $L_{rWGAN}$ also enforces cycle-consistency on feature embeddings between latent space and original visual space.

\subsubsection{Objective} The structure-aware generation can be fused into WGAN and VAEGAN, referred to as SA-WGAN and SA-VAEGAN. Among them, SA-VAEGAN includes an encoder $E(\cdot)$ to map visual instance to a latent variable $z$. The decoder/generator takes $z$ as input and tries to recover the visual feature. Specifically, the VAE objective contains a KL (Kullback-Leibler divergence) and a reconstruction loss. Using MSE (mean square error) as the reconstruction loss, the VAE loss is formulated as
\begin{equation}
\begin{split}
    L_{VAE} &= \text{KL}(E(x_m)||p(z)) + \mathbb{E}[\text{MSE}(G(z, a), x_m)] \\
            & + \mathbb{E}[\text{MSE}(G_2(x_m), x)],
\end{split}
\end{equation}

The overall objective function of SA-VAEGAN in the second stage is defined by
\begin{equation}
  L_{SA-VAEGAN} = L_{VAE} + \gamma_{m} L_{mWGAN} + \gamma_{r}L_{rWGAN},
\end{equation}
where $\gamma_{r}$ is used to balance learning of the second generator and is set to 0.1 in this paper. $\gamma_{m}$ is set to 1. The objective of SA-WGAN contains only last two items.

\begin{table}
  \renewcommand\arraystretch{1.2}
  \caption{Statistics of five benchmark datasets used in the experiments, in
    terms of class embedding dimensions $K_a$, number of seen classes $Y_s$,
    number of unseen classes $Y_u$, number of training samples $X^{tr}$, numbers
    of test seen instances $X^{te}_{s}$ and unseen instances $X^{te}_{u}$.}
  \begin{center}
    \begin{tabularx}{0.47\textwidth}{m{3.5em}
      >{\centering\arraybackslash}X
      >{\centering\arraybackslash}X
      >{\centering\arraybackslash}X
      >{\centering\arraybackslash}m{3.1em}
      >{\centering\arraybackslash}m{2.5em}
      >{\centering\arraybackslash}X}
    \toprule
     Dataset & $K_a$ & $Y_s$ & $Y_u$ & $X^{tr}$ & $X^{te}_{s}$ & $X^{te}_{u}$ \\
    \midrule
     AWA2 & 85    & 40  & 10 & 23,527 & 7,913 & 5,882 \\
     CUB  & 312   & 150 & 50 & 7,075  & 2,967 & 1,764 \\
     SUN  & 102   & 580 & 65 & 14,340 & 1,440 & 2,580 \\
     FLO  & 1,024 & 82  & 20 & 1,640  & 5,394 & 1,155 \\
     \bottomrule
    \end{tabularx}
  \end{center}
  \label{table:dataset}
\end{table}

Give the learned generator $G$, we synthesize unseen instances by inputting
corresponding class embedding $a$ and noise $z$. The generated unseen instances
and real seen instances are then used to learn a linear classifier for final
classification.

In this paper, we use the prototype-instance relationship to characterize the
intrinsic data structure required to preserving during the mapping stage, which
requires us to use the class prototype. In the feature generation stage,
existing ZSL methods learn discriminator by measuring the relationship between
instance, while the prototype enables us to handle structure information which
serves as an additional supervision signal to enhance the discriminator learning
as well as improving the generalizability of the generator. In addition, as
illustrated in Fig. \ref{fig:title} in the paper, compared with other feature structures
such as neighborhood structures, the prototype is easier to compute. Once
obtained, the class prototype can be flexibly applied to mapping sub-network and
GAN sub-network without re-computation in each iteration.

\begin{table*}[h]
  \renewcommand\arraystretch{1.2}
  \caption{Comparing the proposed method with state-of-the-art methods on four benchmarks. We report average per-class top-1 accuracy for unseen (U) classes and seen (S) classes and their harmonic mean (H) in percentage. The best results are highlighted.}
  \begin{center}
    \begin{tabularx}{1.0\textwidth}{m{8em}
        >{\centering\arraybackslash}X
        >{\centering\arraybackslash}X
        >{\centering\arraybackslash}X!{\hspace*{0.8em}}
        >{\centering\arraybackslash}X
        >{\centering\arraybackslash}X
        >{\centering\arraybackslash}X!{\hspace*{0.8em}}
        >{\centering\arraybackslash}X
        >{\centering\arraybackslash}X
        >{\centering\arraybackslash}X!{\hspace*{0.8em}}
        >{\centering\arraybackslash}X
        >{\centering\arraybackslash}X
        >{\centering\arraybackslash}X!{\hspace*{0.8em}}
        >{\centering\arraybackslash}X
        >{\centering\arraybackslash}X
        >{\centering\arraybackslash}X
        }
      \toprule
      \multicolumn{1}{c}{\multirow{3}{*}{Method}} & \multicolumn{3}{c}{\textbf{Zero-Shot Learning}}  & \multicolumn{12}{c}{\textbf{Generalized Zero-Shot Learning}}\\

      \cmidrule(lr{0.8em}){2-4} \cmidrule(lr{0.1em}){5-16}  

        & CUB & SUN & FLO & \multicolumn{3}{c}{AWA2} & \multicolumn{3}{c}{CUB} & \multicolumn{3}{c}{SUN} & \multicolumn{3}{c}{FLO} \\

      \cmidrule(lr{0.8em}){2-4} \cmidrule(lr{1.0em}){5-7} \cmidrule(lr{1.0em}){8-10} \cmidrule(lr{1.0em}){11-13} \cmidrule(lr{0.1em}){14-16} 

       & T1 & T1 & T1 & U & S & H & U & S & H & U & S & H & U & S & H \\ \midrule

    f-CLSWGAN \cite{xian2018feature}             & 57.3 &   -  & 67.2 & -    & -    & -    & 43.7 & 57.7 & 49.7 & 42.6 & 36.6 & 39.4 & 59.0 & 73.8 & 65.6 \\ 
    Cycle-WGAN \cite{felix2018multi}            & 58.6 &   -  & 70.3 & -    & -    & -    & 45.7 & 61.0 & 52.3 & 49.4 & 33.6 & 40.0 & 59.2 & 72.5 & 65.1 \\
    SE-ZSL \cite{kumar2018generalized}         & 59.6 & 63.4 &  -   & 58.3 & 68.1 & 62.8 & 41.5 & 53.3 & 46.7 & 40.9 & 30.5 & 34.9 &  -   &  -   & - \\
    f-VAEGAN \cite{xian2019f}                  & 61.0 & 64.7 & 67.7 & 57.1 & 76.1 & 65.2 & 48.4 & 60.1 & 53.6 & 45.1 & 38.0 & 41.3 & 56.8 & 74.9 & 64.6 \\ 
    SARB-I \cite{paul2019semantically}         & 63.9 & 62.8 &  -   & 30.3 & 93.9 & 46.9 & 55.0 & 58.7 & 56.8 & 50.7 & 35.1 & 41.5 &  -   &   -  & - \\ 
    CADA-VAE \cite{schonfeld2019generalized}   & 60.4 &  -   &  -   & 55.8 & 75.0 & 63.9 & 51.6 & 53.5 & 52.4 & 47.2 & 35.7 & 40.6 &  -   &   -  & - \\
    GMN \cite{sariyildiz2019gradient}          & 64.3 & 63.6 &  -   & -    & -    & -    & 56.1 & 54.3 & 55.2 & 53.2 & 33.0 & 40.7 &  -   &   -  & - \\
    LisGAN \cite{li2019leveraging}             & 58.8 &  -   & 69.6 & -    & -    & -    & 46.5 & 57.9 & 51.6 & 42.9 & 37.8 & 40.2 & 57.7 & 83.8 & 68.3 \\
    ABP \cite{zhu2019learning}                 & 58.5 &  -   &  -   & 55.3 & 72.6 & 62.6 & 47.0 & 54.8 & 50.6 & 45.3 & 36.8 & 40.6 &   -  &  -   & - \\
    TCN \cite{jiang2019transferable}          & 59.5 & 61.5 &  -   & 61.2 & 65.8 & 63.4 & 52.6 & 52.0 & 52.3 & 31.2 & 37.3 & 34.0 &   -  &  -   & - \\
    GXE \cite{li2019rethinking}                & 54.4 & 62.6 &  -   & 56.4 & 81.4 & 66.7 & 47.4 & 47.6 & 47.5 & 36.3 & 42.8 & 39.3 &   -  &  -   & - \\
    RFF \cite{han2020learning}                 &   -  &  -   &      & -    & -    & -    & 52.6 & 56.6 & 54.6 & 45.7 & 38.6 & 41.9 & 65.2 & 78.2 & 71.1\\
    OCD \cite{keshari2020generalized}         & 60.3 & 63.5 &  -   & 59.5 & 73.4 & 65.7 & 44.8 & 59.9 & 51.3 & 44.8 & 42.9 & {43.8} &  -   &   -  & - \\
    DEVB \cite{min2020domain}                  &   -  &   -  &   -  & 63.6 & 70.8 & 67.0 & 53.2 & 60.2 & 56.5 & 45.0 & 37.2 & 40.7 &  -   &   -  & - \\
    TF-VAEGAN \cite{narayan2020latent}         & \textbf{64.9} & \textbf{66.0} & 70.8 & 59.8 & 75.1 & 66.6 & 52.8 & 64.7 & {58.1} & 45.6 & 40.7 & 43.0 & 62.5 & 84.1 & 71.7 \\
    \midrule
      SA-WGAN                              & 62.4 & 63.7 & 80.9 & 59.7 & 73.2 & 65.8 & 53.8 & 61.6 & 57.4 & 47.6 & 40.9 & 44.0 & 77.1 & 88.4 & {82.4} \\
      SA-VAEGAN                             & 64.5 & 64.4 & \textbf{89.3} & 59.4 & 79.5 & \textbf{68.0} & 55.7 & 62.0 & \textbf{58.7} & 51.3 & 39.1 & \textbf{44.4}  & 85.0 & 87.7 & \textbf{86.4} \\
    \bottomrule
    \end{tabularx}
  \end{center}
  \label{table:GZSL}
\end{table*}

\section{Experiments}
In this section, we first describe the benchmarks, implementation details, and
evaluation protocols. We then report the results of our method based on both
WGAN and VAEGAN, and compare them with state-of-the-art ZSL/GZSL approaches.
After that, we conduct ablation studies to investigate the effectiveness of our
method.

\subsection{Dataset}
We validate the proposed method on four commonly used ZSL/GZSL datasets, which
are Animals with Attributes 2 (AWA2) \cite{xian2018zero}, CUB-200-2011 (CUB)
\cite{wah2011caltech}, SUN with Attributes (SUN) \cite{patterson2012sun}, Oxford
Flowers (FLO) \cite{nilsback2008automated}. Among them, CUB and FLO are
fine-grained datasets, and the other two are coarse-grained datasets. The
semantic labels used by the datasets are provided by the category attributes.
For fair comparison against existing approaches, the split of seen/unseen
classes follows the newly proposed setting (PS) \cite{xian2018zero}, which
strictly separates the training classes from the pre-trained model. For FLO, we
use the standard split and sentence-based semantic descriptions provided by Reed
\textit{et al.} \cite{reed2016learning}. We list the statistics of these
datasets in Table \ref{table:dataset}.

\begin{table}[h]
  \renewcommand\arraystretch{1.2}
  \caption{Classification accuracy (\%) of conventional zero-shot learning for standard split (SS). The best results are highlighted.}
  \begin{center}
    \begin{tabularx}{0.46\textwidth}{m{7em}
      >{\centering\arraybackslash}X
      >{\centering\arraybackslash}X
      >{\centering\arraybackslash}X}
    \toprule
     Method                              & CUB  & AWA2 & SUN   \\
    \midrule
     CONSE\cite{norouzi2013zero}         & 36.7 & 67.9 & 44.2  \\
     SSE \cite{zhang2016learning}        & 43.7 & 67.5 & 54.5  \\
     LATEM \cite{xian2016latent}         & 49.4 & 68.7 & 56.9  \\
     ALE\cite{akata2013label}            & 53.2 & 80.3 & 59.1  \\
     DEVISE\cite{frome2013devise}        & 53.2 & 68.6 & 57.5  \\
     SJE\cite{akata2015evaluation}       & 55.3 & 69.5 & 57.1  \\
     ESZSL\cite{romera2015embarrassingly}& 55.1 & 75.6 & 57.3  \\
     SYNC\cite{changpinyo2016synthesized}& 54.1 & 71.2 & 59.1  \\
     SAE\cite{kodirov2017semantic}       & 33.4 & 80.2 & 42.4  \\
     GFZSL\cite{verma2017simple}         & 53.0 & 79.3 & 62.9  \\
     SE-ZSL\cite{kumar2018generalized}   & 60.3 & 80.8 & 64.5  \\
     DCN\cite{liu2018generalized}        & 55.6 &  -   & {67.4}  \\
     OCD\cite{keshari2020generalized}    & {60.8} & {81.7} & \textbf{68.9}  \\
    \midrule
     SA-WGAN      & {66.7} & {82.1} & {66.6} \\
     SA-VAEGAN     & \textbf{67.5} & \textbf{82.2} & {67.1} \\
    \bottomrule
    \end{tabularx}
  \end{center}
  \label{table:ZSL-ss}
\end{table}

\subsection{Implementation Details}
Throughout the paper, visual features are extracted from the image with 2048-dim
using ResNet-101 \cite{he2016deep} extractor. There is no pre-processing
technique used in the feature extraction. Specifically, the ResNet-101 model is
pre-trained on ImageNet 1K without fine-tuning. Similar to
\cite{xian2018feature, felix2018multi}, for FLO, we extract 1024-dim
character-based CNN-RNN \cite{reed2016learning} features that encode the text
description of the image containing the fine-grained visual descriptions (10
sentences per image). There is no interaction between seen and unseen sentences
during CNN-RNN training. In this case, the per-class sentences are build by
averaging the CNN-RNN features of the same class. Note that for FLO, all methods
except DeViSE \cite{frome2013devise} use the RNN-CNN description
\cite{reed2016learning}.

We have formally defined the objective function and depicted the framework in
Fig.~\ref{fig:framework}. The encoder $E$, generator $G$, $G_2$, discriminator
$D$, $D_2$ are implemented as multiple-layer perceptron (MLP) with a single
hidden layer containing 4096 nodes and LeakyReLU \cite{maas2013rectifier} as the
activation function. We use Adam optimizer with $\beta_1=0.5$, $\beta_2=0.999$
for both stages. The batch sizes for mapping network and GAN training are 256
and 64. We train 100 epochs in the mapping sub-network and 1000 epochs for the
learning of feature generation. For all datasets, we uniformly set the learning
rate to 1e-4.

\subsection{Evaluation Protocols}
For evaluation protocols, we follow the protocol proposed in \cite{xian2018zero}
to evaluate ZSL and GZSL performance. In particular, we average the correct
predictions for each class and report the top-1 accuracy of per class. For
conventional ZSL setting, the evaluation metric is defined as follows, 
\begin{equation}
MCA_{u} = \frac{1}{|U|}\sum_{y^u\in{U}}acc_{y^u}
\end{equation}
where $acc_{y^u}$ denotes the top-1 accuracy on the unseen test data for each
class in the unseen domain $U$. The $|U|$ denotes the class number.

In the generalized ZSL setting, we compute the averaget per class top accuracy
for both seen and unseen classes and report the harmonic mean of seen and unseen
accuracy. The GZSL evaluation protocal is computed as follows,
\begin{equation}
H=\frac{2*MCA_s*MCA_u}{MCA_s+MCA_u}
\end{equation}
where $MCA_u$,$MCA_s$ denote the mean class accuracy on unseen and seen classes
respectively. A well-trained GZSL model is required to reach good results on the
unseen domain while maintaining the recognition capability in the seen domain. A
higher harmonic mean indicates a better balance between seen classes and unseen
classes.

\subsection{Comparing with the State-of-the-Art}

\subsubsection{Conventional ZSL Using Standard Splits} Table \ref{table:ZSL-ss} summarizes the
results of conventional zero-shot learning on AWA2, CUB and SUN datasets. These
results are obtained from Xian \textit{et al.} \cite{xian2017zero, xian2018zero}
using standard split (SS) with ResNet-101 as the feature extractor. From the
table, we can see that the classification accuracy obtained on the SS protocol
on AWA2, CUB, and SUN datasets are 82.2\%, 67.5\%, 67.1\%, respectively. SA-GAN
does not outperforms all the prior approaches on all dataset for the inductive
setting. However, compared with the best existing methods, our method
illustrates a significant advantage on the CUB dataset with around 7\%
improvement. In addition, we achieve consistent improvement on the AWA2 dataset
with 0.5\% improvement. The result on SUN is also competitive with an accuracy
of 67.1\%. Although OCD obtains better results on SUN, the results of SA-GAN on
the other two datasets are higher with improvents of 6.7\% and 0.5\%
respectively. We attribute this performance gain to the increased discriminitive
power by the mapping sub-nework with .

\begin{figure*}[ht]
  \begin{center}
    \includegraphics[width=1.0\linewidth]{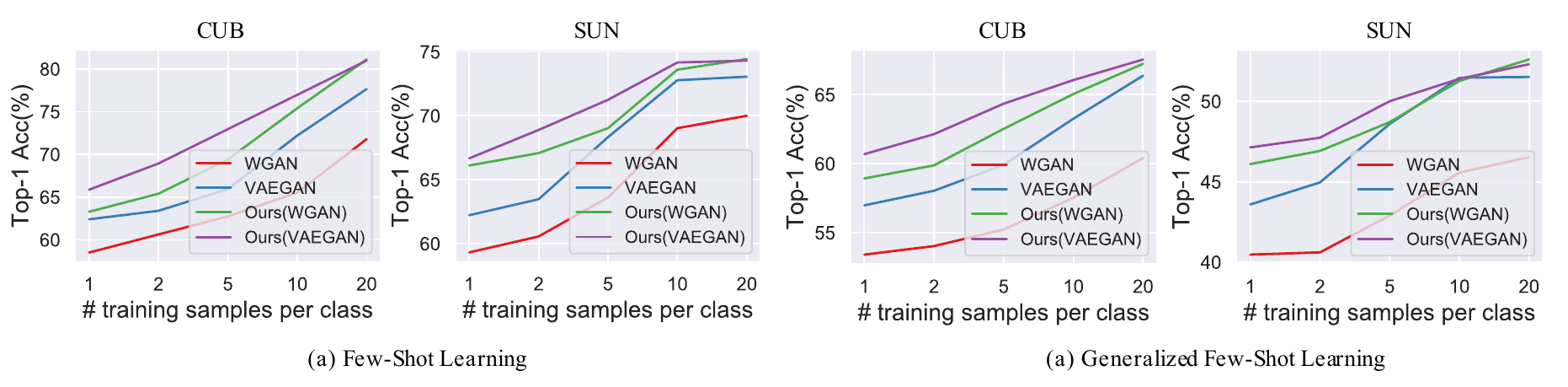}
  \end{center}
  \caption{FSL and GFSL results on CUB and SUN with increasing number of training samples per novel class.}
  \label{fig:few-shot}
\end{figure*}

\begin{table*}
  \renewcommand\arraystretch{1.2}
  \caption{Zero-shot learning using fine-tuned feature on CUB, AWA2 and SUN datasets.}
  \begin{center}
    \begin{tabularx}{0.95\textwidth}{m{8em}
        >{\centering\arraybackslash}X
        >{\centering\arraybackslash}X
        >{\centering\arraybackslash}X!{\hspace*{0.8em}}
        >{\centering\arraybackslash}X
        >{\centering\arraybackslash}X
        >{\centering\arraybackslash}X!{\hspace*{0.8em}}
        >{\centering\arraybackslash}X
        >{\centering\arraybackslash}X
        >{\centering\arraybackslash}X!{\hspace*{0.8em}}
        >{\centering\arraybackslash}X
        >{\centering\arraybackslash}X
        >{\centering\arraybackslash}X
        }
      \toprule
      \multicolumn{1}{c}{\multirow{3}{*}{Method}} & \multicolumn{3}{c}{\textbf{Zero-Shot Learning}}  & \multicolumn{9}{c}{\textbf{Generalized Zero-Shot Learning}}\\

      \cmidrule(lr{0.8em}){2-4} \cmidrule(lr{0.1em}){5-13}  

       & AWA2 & CUB & SUN &  \multicolumn{3}{c}{AWA2} & \multicolumn{3}{c}{CUB} & \multicolumn{3}{c}{SUN}  \\

      \cmidrule(lr{0.8em}){2-4} \cmidrule(lr{0.8em}){5-7} \cmidrule(lr{0.8em}){8-10} \cmidrule(lr{0.1em}){11-13} 

      &T1 & T1 & T1 & U & S & H & U & S & H & U & S & H  \\ \midrule

    SBAR-I \cite{paul2019semantically}  & 65.2 & 63.9 & 62.8 & 30.3 & 93.9 & 46.9 & 55.0 & 58.7 & 56.8 & 50.7 & 35.1 & 41.5  \\
    f-VAEGAN \cite{xian2019f}           & 70.3 & 72.9 & 65.6 & 57.1 & 76.1 & 65.2 & 63.2 & 75.6 & 68.9 & 50.1 & 37.8 & 43.1  \\
    TF-VAEGAN \cite{narayan2020latent}  & 73.4 & 74.3 & 66.7 & 55.5 & 83.6 & 66.7 & 63.8 & 79.3 & \textbf{70.7} & 41.8 & 51.9 & {46.3}  \\
    \midrule
    {SA-WGAN}    & 68.1 & 71.3 & 63.5 & 80.6 & 62.1 & {70.2} & 73.7 & 63.5 & 68.2 & 43.3 & 49.5 & 46.2  \\
    {SA-VAEGAN}  & 70.5 & 72.3 & 63.5 & 82.3 & 63.9 & \textbf{71.9} & 75.7 & 65.4 & \textcolor{blue}{70.2} & 49.9 & 43.7 & \textbf{46.7}  \\
    \bottomrule
     
    \end{tabularx}
  \end{center}
  \label{table:GZSL-ft}
\end{table*}

\subsubsection{ZSL/GZSL Setting using Proposed Splits} Table \ref{table:GZSL} compares our method with
state-of-art methods proposed recently on four zero-shot learning datasets under
both ZSL and GZSL settings using proposed split (PS) \textit{et al.} without
fine-tuning the feature extractor on the target datasets. \cite{xian2017zero,
  xian2018zero}. SA-GAN improves over f-CLSWGAN \cite{xian2018feature},
Cycle-WGAN \cite{felix2018multi}, SE-ZSL \cite{kumar2018generalized}, f-VAEGAN
\cite{xian2019f}, SARB-I \cite{paul2019semantically}, CADA-VAE
\cite{schonfeld2019generalized}, GMN \cite{sariyildiz2019gradient}, LisGAN
\cite{li2019leveraging}, ABP \cite{zhu2019learning}, TCN
\cite{jiang2019transferable}, GXE \cite{li2019rethinking}, RFF
\cite{han2020learning}, OCD \cite{keshari2020generalized}, DEVB
\cite{min2020domain} and TF-VAEGAN \cite{narayan2020latent} over all the
benchmarks measured by harmonic mean. Those prior methods utilized the semantic
space for class embedding without considering the topological structure in the
sample generation and thus restricted by the generalization problem. We observe
that SA-GAN achieves new state-of-art results, i.e., 68.0\% on AWA2, 58.7\% on
CUB, 44.4\% on SUN, and 86.4\% on FLO, which are 4.5\%, 5.1\%, 3.1\%, and 21.8\%
improvements from VAEGAN baseline. This is because SA-GAN not only promotes the
feature discriminative power in the mapping space like SABR-I but also preserves
the hidden structure information characterized by the instance-prototype
relationship in the mapping stage and the feature generation stage. A similar
result can be seen when we compare SA-GAN with f-CLSWGAN. Concretely, SA-GAN
improves the results by 7.7\%, 5.0\%, and 20.8\% on CUB, SUN, and FLO
respectively. Note that SA-GAN using a mapping subnetwork in a two-stage scheme
like SABR-I, but obtains better results improving the results by 21.1\%, 1.9\%,
2.9\% on AWA2, CUB, and SUN. This is because SABR-I uses an additional mapping
sub-subnetwork to address the hubness problem. However, projecting the visual
feature by the classifier and semantic alignment reduces the bias of the seen
classes at the cost of deconstructing the structure information embedded in the
visual space.

Most existing G/ZSL approaches performs well on the seen domain but fail to
achieve satisfying results for unseen classes, even through they are try to
addres the isuue, which indicates that a large biases towards seen classes is
challenging in the ZSL field. Our model mitigate the gap between seen and unseen
classes, as shown in Table \ref{table:GZSL}. With the structure-preserved
mapping and structure-aware generator, the $MCA_u$ increases and a better
balance between the accuracy of seen and unseen classes is obtained. Therefore,
our generative-based approaches competitively benefit the zero-shot learning in
the realistic and challenging task. This also partially explain why there are a
few unpromising result in the Table \ref{table:GZSL} and Table
\ref{table:ZSL-ss}. Although not state-of-art on the unseen split, our method
cares more about the overall performance where both seen and unseen classes are
included for prediction.

\subsection{(Generalized)~Few-shot Learning}


In few-shot learning, the data classes are split into base classes containing a
large number of training data and novel classes that contain few labeled
examples each class. A good few-shot model is required to achieve good results
on novel classes. Similar to GZSL, the generalized few-shot learning measure the
performance across all classes. Following f-VAEGAN, we use the same split as
(G)ZSL, i.e. 150 base classes and 50 novel classes on CUB. However, there is no
training data available for novel classes in ZSL, since all data of novel
classes from the original dataset are combined for ZSL evaluation. To enable
few-shot learning, we split the novel data into training/test according to the
protocol in the original dataset. We use the base class data as the support set.
During training, we randomly sample (1, 2, 5, 10, 20) instances, and combine
them with support data for model learning. Other settings remain the same as
(G)ZSL. In the test time, we report top-1 accuracy in both few-shot and
generalized few-shot learning.

\begin{table*}
  \renewcommand\arraystretch{1.2}
  \caption{Contribution of components to the generalized zeroshot learning (H)
    on the CUB dataset. SP-Map denotes structure-preserving mapping, mWGAN
    denotes structure-aware feature generation in the mapped space, and rWGAN
    reconstructs original visual features from latent ones.}
  \begin{center}
    \begin{tabularx}{0.95\textwidth}{
      >{\centering\arraybackslash}X
      >{\centering\arraybackslash}X
      >{\centering\arraybackslash}X
      >{\centering\arraybackslash}X
      >{\centering\arraybackslash}X
      >{\centering\arraybackslash}X
      >{\centering\arraybackslash}X
      >{\centering\arraybackslash}X}
    \toprule
      Baseline &  SP-MAP     &  mWGAN    &  rWGAN      & ZSL   &   U   &   S   &   H  \\
    \midrule
      \checkmark &             &            &            & 56.81 & 48.15 & 56.68 & 52.06 \\
      \checkmark & \checkmark  &            &            & 62.03 & 52.89 & 60.26 & 56.34 \\
      \checkmark &             & \checkmark &            & 63.05 & 53.51 & 60.51 & 56.80 \\
      \checkmark & \checkmark  & \checkmark &            & 64.04 & 54.04 & 62.99 & 58.17 \\
      \checkmark & \checkmark  & \checkmark & \checkmark & 64.48 & 56.39 & 61.31 & 58.74 \\
    \bottomrule
    \end{tabularx}
  \end{center}
  \label{table:component}
\end{table*}

Fig.~\ref{fig:few-shot} shows that the results obtained by our method
significantly improves over the GZSL setting even only using a few labeled real
instances. Specifically, compared with ZSL, one-shot learning uses one labelled
real instance for each novel class and improve the accuracy by 10\%. The
accuracy improvements from 1 to 20 shots is over 15\%. Besides, while the top-1
accuracy increases with the number of shots for all compared methods, our method
outperforms the baseline by a large margin in both few-shot and generalized
few-shot learning, which indicates a strong generalization ability of our
method. 

\subsection{Ablation Studies}
In this section, we study the properties of the proposed method by conducting a
series of ablation experiments, including the target feature dimension, number
of synthetic samples.

\begin{figure}
  \begin{center}
    \includegraphics[width=0.99\linewidth]{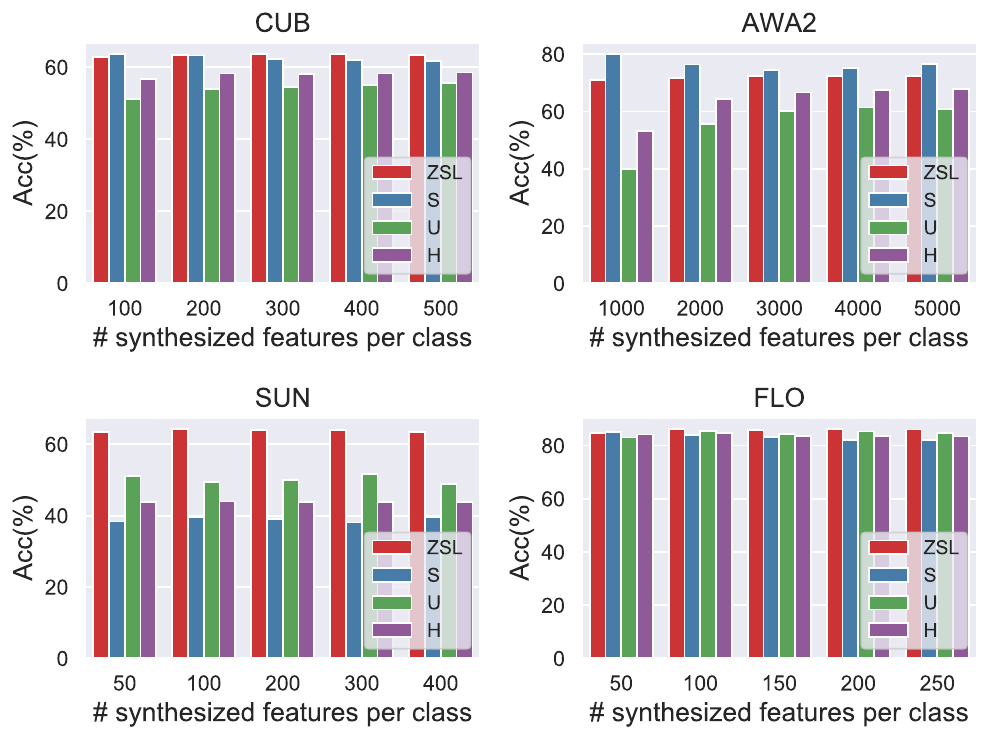}
  \end{center}
  \caption{Impact of the number of synthetic instances on the CUB dataset.}
  \label{fig:syn-num}
\end{figure}

\begin{figure}
  \begin{center}
    \includegraphics[width=0.99\linewidth]{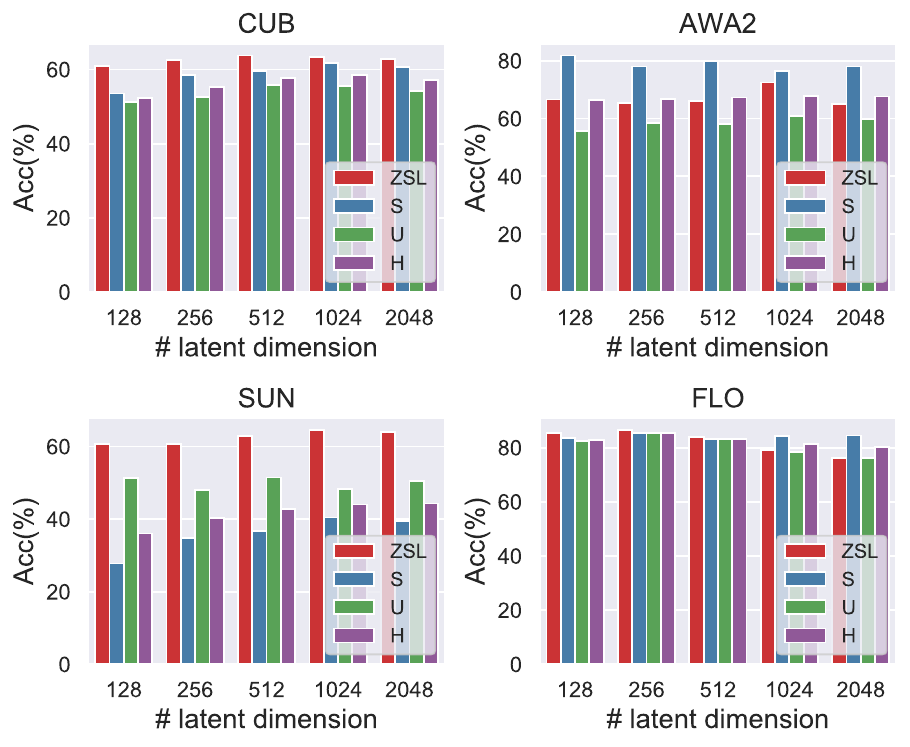}
  \end{center}
  \caption{Impact of the latent dimension in terms of ZSL, U, S, H on four  datasets.}
  \label{fig:dim}
\end{figure}

\begin{figure*}
  \begin{center}
    \includegraphics[width=0.95\linewidth]{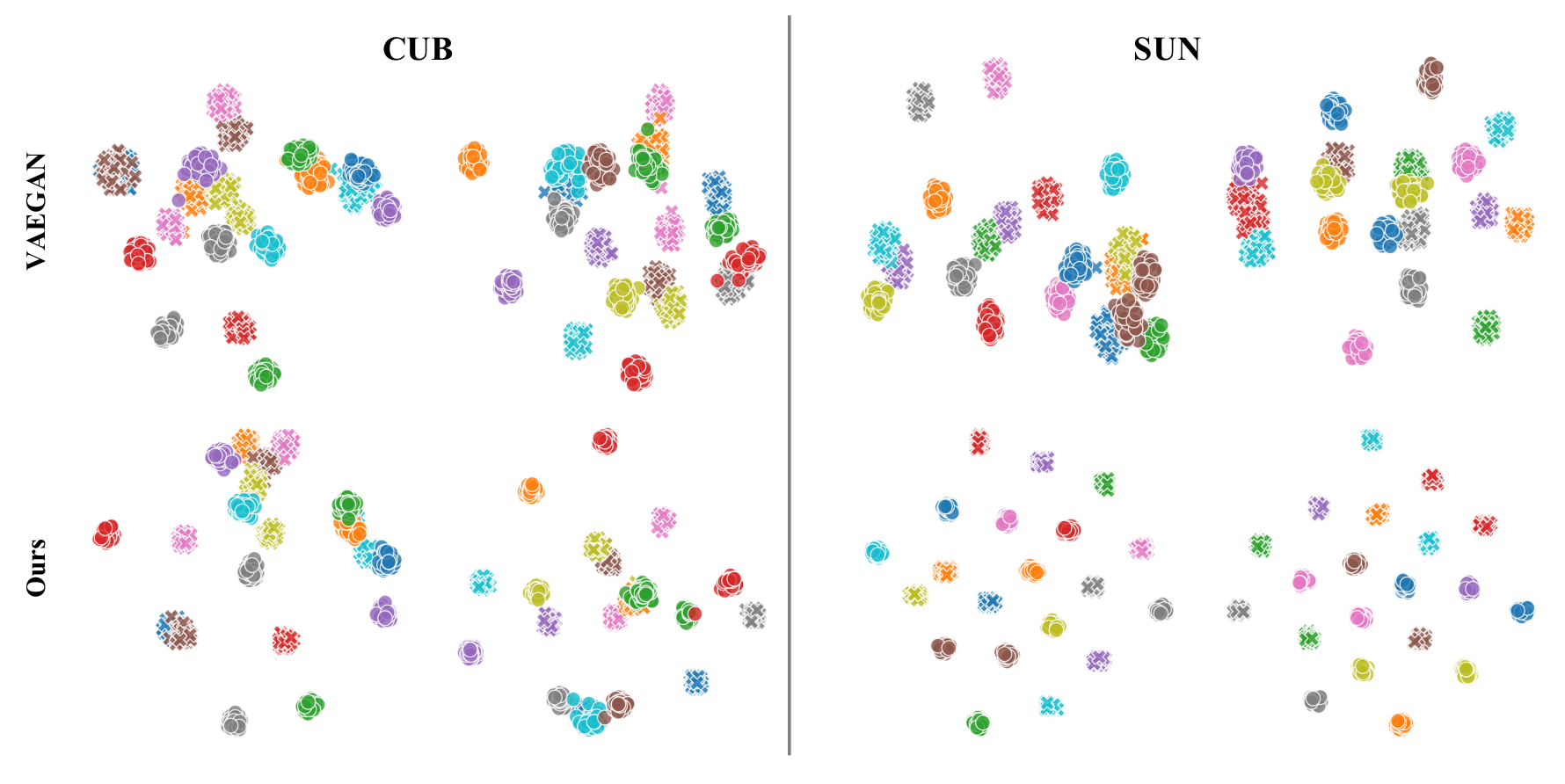}
  \end{center}
  \caption{Comparing VAEGAN and SA-VAEGAN using t-SNE embeddings of the
    generated feature on CUB and SUN. The top row illustrates VAEGAN, and the
    bottom row shows our method. The symbol $\bullet$ denotes the instance of
    seen classes, and $\mathbf{\times}$ denotes the instance of unseen classes.}
  \label{fig:tsne-cub}
\end{figure*}

\subsubsection{ZSL on Fine-Tuned Feature}
We also evaluate our method using fine-tuned features on three datasets in
Fig.~\ref{table:GZSL-ft}. Overall, the improvement is significant when compared
with approaches without using fine-tuned features. Compared with the methods
using the fine-tuned features, our method also illustrates the consistent
improvement in GZSL on two datasets.

\subsubsection{Metric of Instance-Prototye Measurement}
Note that, throughout the papaer we use L2 distance as the metric to evaluate
the relationship between instance and its class prototypes. Here we conduct
experiments to test the model performace under different distnace metrics
including L1, L2, and cosine distance and record the results in Table
\ref{table:dist-metric} The results show that L2 distance is more effective for
the model to achieve a higher performance.

\begin{table}
  \renewcommand\arraystretch{1.2}
  \caption{Results comparison of different distance metrics on the cub dataset.}
  \begin{center}
    \begin{tabularx}{0.45\textwidth}{m{3.5em}
      >{\centering\arraybackslash}X
      >{\centering\arraybackslash}X
      >{\centering\arraybackslash}X
      >{\centering\arraybackslash}X}
      \toprule
      & ZSL & U & S & H\\
      \midrule
      L1      & 62.81 & 52.04 & 60.64 & 56.01\\
      Cosine  & 62.77 & 54.30 & 61.95 & 57.87\\
      L2      & 64.54 & 56.53 & 61.41 & 58.87\\
      \bottomrule
    \end{tabularx}
  \end{center}
  \label{table:dist-metric}
\end{table}

\subsubsection{Impact of Model Components}

SA-GAN has utilized multiple techniques to improve the performance. We evaluate
the effectiveness of components used in SA-VAEGAN. Table \ref{table:component}
summarizes the results of four settings. The baseline model consists of a
generator $G$, a discriminator $D$, and an encoder $E$. Based on the baseline,
we evaluate the model performance by introducing structure-preserving mapping
(SP-Map), mapped GAN (mWGAN), and reconstructed GAN (rWGAN). We report harmonic
mean (H) on the CUB dataset. Table \ref{table:component} shows significant
improvement over the VAEGAN method. The complete version gives the highest
results, achieving a whopping accuracy gain of 6.6\%. Specifically, the
introduced structure-preserving mapping remarkably enhances the harmonic mean by
a large margin (4.7\%). This is because the SP-Map not only promotes feature
discriminative power but retains prototype structure for high-quality mapping.
After utilizing the mWGAN, the model performance takes a further enhancement.
The baseline discriminator measures the relation of an instance and its class
embedding but mWGAN further includes prototypes, which considers more structural
information. By combining these modules, our method yields the best results.

Table \ref{table:ab-sp} further shows the result comparison between using and
without using structure-preserving loss. From the table, we can see that the
structure-preserving loss improves the model performance in terms of all
metrics, which confirms the effectiveness of structure-preserving loss.

\begin{table}[h]
  \renewcommand\arraystretch{1.2}
  \caption{Results comparison with/without Structure-Preserving (SP) loss on the CUB dataset.}
  \begin{center}
    \begin{tabularx}{0.47\textwidth}{m{5em}
      >{\centering\arraybackslash}X
      >{\centering\arraybackslash}X
      >{\centering\arraybackslash}X
      >{\centering\arraybackslash}X}
      \toprule
      & ZSL & U & S & H\\
      \midrule
      without SP   & 63.27 & 55.82 & 58.79 & 57.26 \\
      with SP      & 64.54 & 56.53 & 61.41 & 58.87 \\
      \bottomrule
    \end{tabularx}
  \end{center}
  \label{table:ab-sp}
\end{table}

Note that the visual features are extracted by the ImageNet pre-trained network.
The extracted feature with a high dimension contains redundant information. By
using a mapping network, we can project the visual feature to a compact space to
reduce the influence of redundant information. The mapping network with
pre-defined constraints can also endow the mapped features with new properties
such as increased discriminative power by the classification loss and center
loss constraint, making the new features more task-specific. In this paper, we
also encode structural information within the mapped feature by the proposed
structure-preserving constraint. Our experiments show that the mapping
sub-network improves the model generalizability and benefit the zero-shot
recognition.

Existing works enforce a cycle-consistency between visual features and semantic
embeddings. The proposed reconstructed GAN enables a cycle consistency between
the mapping space and the original visual space. In the ablation study, we
verify the effectiveness of the reconstructed GAN in Table \ref{table:component} in
the paper. More detailed results are also illustrated in Table 1 in the
supplementary materials. These experiments show that the reconstructed GAN
further improves the model performance by over 0.5\%.

\subsubsection{Impact of Synthetic Numbers}

We show in Fig.~\ref{fig:syn-num} results of our method under different numbers
of synthesized instances of unseen classes on four datasets. In general, the
performance remains stable across benchmarks. When the amount of synthesized
instances is small, unseen classes' accuracy is low (ZSL, U, H). This is because
insufficient unseen class data causes data imbalance problem, encouraging the
GZSL classifier to be more focused on learning from seen class data. The issue
is alleviated when the amount increases. As a result, the ZSL, U, and H results
improve rapidly. Besides, to address the imbalanced problem, the fine-grained
dataset usually required a smaller number of synthetic instances before
achieving the best performance, while for coarse-grained dataset, the optimal
number is larger.

\subsubsection{Impact of the Latent Dimension}

The effect of the latent dimension of the hidden feature, evaluated by the
proposed method, is illustrated in Fig.~\ref{fig:dim}. In general, when the
dimension is low, the performance is poor. One reason is that the very low
feature dimension is not enough to fully describe the target class. As the
dimension increases, the unseen accuracy and harmonic mean gain consistent
improvement, and when the dimension exceeds a certain threshold, the performance
does not improve. We use 1,024 as the mapped dimension throughout this paper
because our method has already achieved good results on the selected datasets.

\subsubsection{Feature Visualization}

\begin{figure}
  \begin{center}
    \includegraphics[width=0.99\linewidth]{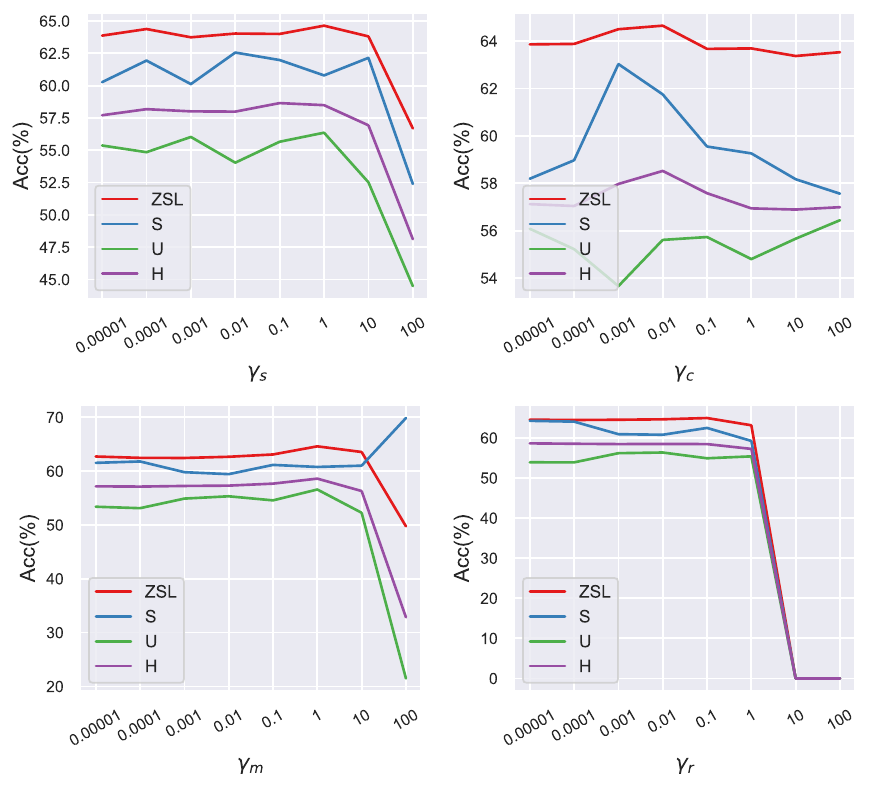}
  \end{center}
  \caption{Model performances on the CUB dataset with different coefficients}
  \label{fig:weight-analysis}
\end{figure}

In Fig.~\ref{fig:tsne-cub}, we provide statistics to
compare SA-VAEGAN with VAEGAN in terms of the synthetic features on the CUB
dataset. We selected the ten unseen classes and ten seen classes for T-SNE
\cite{maaten2008visualizing} visualization. As shown in Fig.~\ref{fig:tsne-cub},
there are overlapping among unseen classes for VAEGAN. However, our method leads
to more visible clusters, where the synthesized samples of the selected category
become more compact. The synthesized features are more discriminative, and those
overlapped classes become more separable. This proves that the SA-GAN is
powerful in generating more discriminative features for unseen classes.

\subsection{Analysis of Loss Coefficients}
We study the effect of loss coefficients $\gamma_s$, $\gamma_c$, $\gamma_{w}$,
$\gamma_r$ to obtain a more intuitive observation on the module influence in
Fig.~\ref{fig:weight-analysis}. These experiments are all conducted on the CUB
dataset. $\gamma_s$ controls the importance of structure-preserving, $\gamma_c$
controls the importance of features' discriminative power, $\gamma_m$ controls
the importance of GAN, while $\gamma_r$ controls the importance of visual
feature recovery. These figures show that the model performance remains stable
when the loss coefficients $\gamma_s$, $\gamma_{m}$, $\gamma_r$ are small.
However, the performance decreases dramatically when these coefficients become
very large. Besides, different coefficients have different best values. For
$\gamma_c$, the harmonic mean achieves the best performance at 0.01. This is
because the center loss can promote the features' discriminative power, and when
$\gamma_c$ is low, the features are not discriminative enough to benefit model
performance. However, when the $\gamma_c$ is too large, the model will be prone
to emphasizing the classification on the seen classes without considering the
importance of structural integrity, thus deteriorating the performance on unseen
classes.

\section{Conclusion}
We propose in this paper a novel ZSL scheme, termed SA-GAN, that explicitly
accounts for the topological structure of samples throughout the ZSL pipeline.
This is accomplished through injecting a constraint loss to preserve the initial
geometric structure when learning a discriminative latent space, followed by
conducting GAN training under additional supervision from a structure-aware
discriminator and a reconstruction module. The former guides the generator to
produce discriminative and structure-aware instances, while the latter enforces
the consistency between the original visual space and the GAN space.
Experimental results on four benchmarks showcase that, SA-GAN consistently
outperforms the state of the art, indicating that SA-GAN generates high-quality
instances and enjoys a good generalization ability.

\bibliography{egbib}
\bibliographystyle{IEEEtran}

\vfill
\end{document}